%% file: main.tex
\documentclass[11pt,a4paper]{article}
\usepackage[hyperref]{emnlp-ijcnlp-2019}
\usepackage{times}
\usepackage{latexsym}
\usepackage{graphicx}
\usepackage{amsmath}
\usepackage{xspace}
\usepackage{booktabs}
\usepackage{todonotes}
\usepackage{comment}

\input{math_commands.tex}

\newcommand\wtq{{\sc WikiTableQuestions}\xspace}
\newcommand\swtq{{\sc WTQ}\xspace}
\newcommand\wsq{{\sc WikiSQL}\xspace}
\newcommand{\db}[1]{[[{#1}]]}
\DeclareMathOperator*{\LSTM}{LSTM}
\DeclareMathOperator*{\att}{Attention}

\usepackage{url}

\aclfinalcopy 

\title{Learning Semantic Parsers from Denotations with Latent \\ Structured Alignments and Abstract Programs}

\author{Bailin Wang, Ivan Titov \and Mirella Lapata \\
  Institute for Language, Cognition and Computation \\
  School of Informatics, University of Edinburgh \\
  {\tt bailin.wang@ed.ac.uk, \tt \{ititov,mlap\}@inf.ed.ac.uk}}

\date{}

\begin{document}
\maketitle
\begin{abstract}

  Semantic parsing aims to map natural language utterances onto
  machine interpretable meaning representations, aka programs whose
  execution against a real-world environment produces a denotation.
  Weakly-supervised semantic parsers are trained on
  utterance-denotation pairs treating programs as latent. The task is
  challenging due to the large search space and spuriousness of
  programs which may execute to the correct answer but do not
  generalize to unseen examples.  Our goal is to instill an inductive
  bias in the parser to help it distinguish between spurious and
  correct programs.  We capitalize on the intuition that correct
  programs would likely respect certain structural constraints were
  they to be aligned to the question (e.g., program fragments are
  unlikely to align to overlapping text spans) and propose to model
  alignments as structured latent variables.  In order to make the
  latent-alignment framework tractable, we decompose the parsing task
  into (1) predicting a partial ``abstract program'' and (2) refining
  it while modeling structured alignments with differential dynamic
  programming. We obtain state-of-the-art performance on the \wtq and
  \wsq datasets. When compared to a standard attention baseline, we
  observe that the proposed structured-alignment mechanism is highly
  beneficial.
\end{abstract}

 \section{Introduction}
\input{intro}

\section{Background}
\input{backgroud}

\section{Model}
\input{model}

\section{Experiments}
\input{experiment}

\section{Related Work}
\input{related_work}

\section{Conclusions}

\input{conclusion}

\section*{Acknowledgements}
We would like to thank the anonymous reviewers for their valuable comments. We
gratefully acknowledge the support of the European Research Council (Titov: ERC StG BroadSem 678254; Lapata: ERC CoG TransModal 681760) and the Dutch National Science Foundation (NWO VIDI 639.022.518).

\bibliography{main}
\bibliographystyle{acl_natbib}

\appendix
\input{appendix}

\end{document}

%% file: math_commands.tex
\usepackage{amsmath,amsfonts,bm}

\def\eqref#1{equation~\ref{#1}}

\def\1{\bm{1}}

\def\va{{\bm{a}}}
\def\vb{{\bm{b}}}
\def\vc{{\bm{c}}}

\def\vg{{\bm{g}}}

\def\vl{{\bm{l}}}

\def\vr{{\bm{r}}}
\def\vs{{\bm{s}}}

\def\mA{{\bm{A}}}

\def\mM{{\bm{M}}}

\DeclareMathAlphabet{\mathsfit}{\encodingdefault}{\sfdefault}{m}{sl}
\SetMathAlphabet{\mathsfit}{bold}{\encodingdefault}{\sfdefault}{bx}{n}

\newcommand{\E}{\mathbb{E}}

\newcommand{\R}{\mathbb{R}}

\newcommand{\softmax}{\mathrm{softmax}}

\DeclareMathOperator*{\argmax}{arg\,max}

%% file: intro.tex
\begin{figure}[t]
\vspace{-3mm}
\centering
\includegraphics[width=0.45 \textwidth]{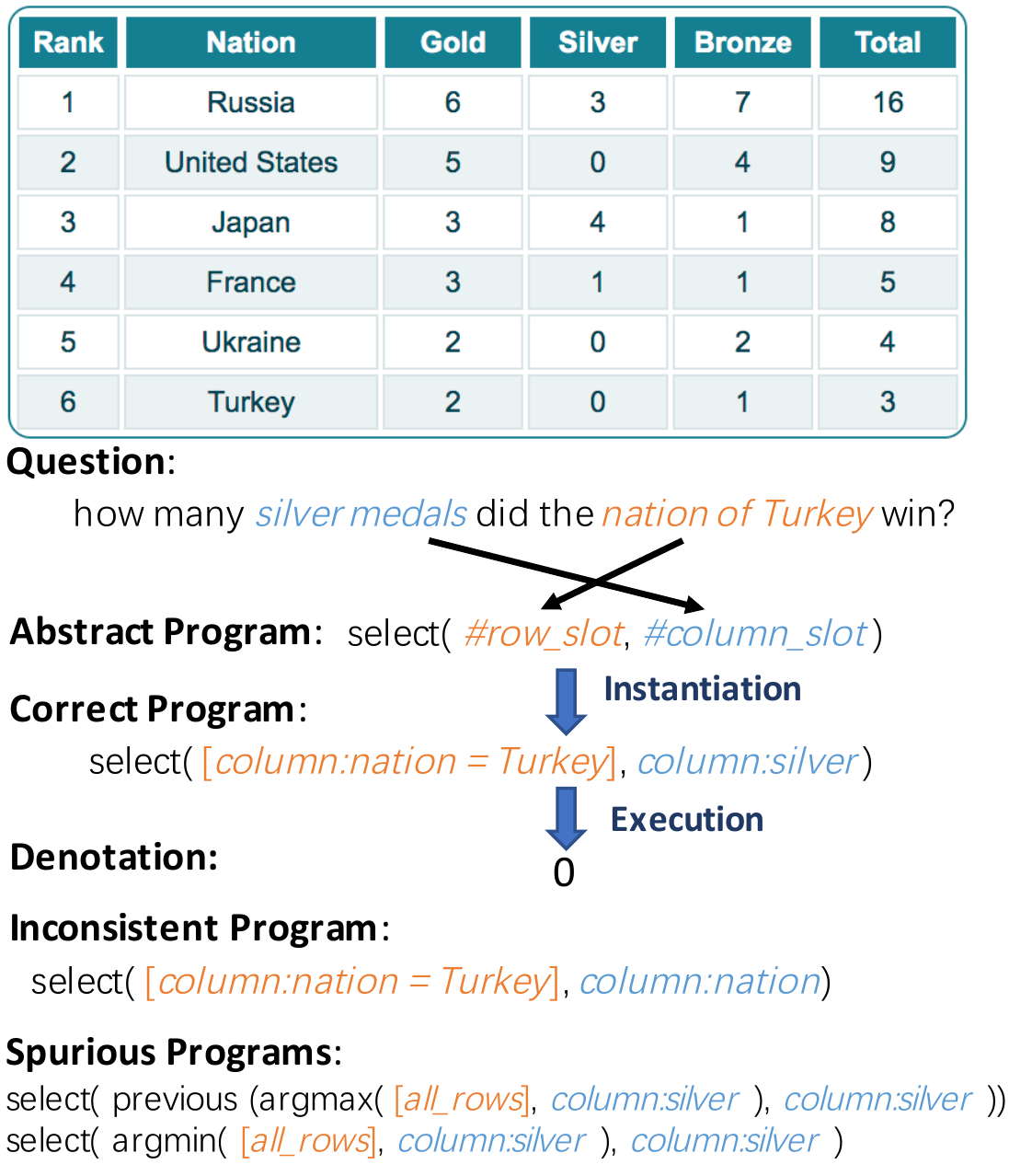}
\vspace{-2mm}
\caption{After generating an abstract program for a question, our parser finds alignments between slots (with prefix \#) and question spans.
Based on the alignment, it instantiates each slot and a complete
program is executed against a table to obtain a denotation.
}
\label{fig:example}
\end{figure}

Semantic parsing is the task of translating natural language to
machine interpretable meaning representations.  Typically, it
requires mapping a natural language utterance onto a program, which is
executed against a knowledge base to obtain an answer or a denotation.
Most previous work
\cite{zettlemoyer2005learning,wong2007learning,lu2008generative,jia2016data}
has focused on the supervised setting where a model is learned from
question-program pairs.  Weakly supervised semantic parsing
\cite{berant2013semantic,liang11dcs} reduces the burden of annotating
programs by learning from questions paired with their answers (or
denotations).

Two major challenges arise when learning from denotations: 1) training
of the semantic parser requires exploring a large search space of
possible programs to find those which are \textit{consistent}, and
execute to correct denotations; 2) the parser should be robust to
\textit{spurious} programs which accidentally execute to correct
denotations, but do not reflect the semantics of the question.
In this paper, we propose a weakly-supervised neural semantic parser
that features structured latent alignments to bias learning towards
\textit{correct} programs which are consistent but not spurious.

Our intuition is that correct programs should respect certain
constraints were they to be aligned to the question text, while
spurious and inconsistent programs do not.  For instance, in
Figure~\ref{fig:example}, the answer to the question (``0") can be
obtained by executing the correct program which selects the number of
Turkey's silver medals.  However, the same answer can be also obtained
by the spurious programs shown in the figure.\footnote{The first program
  can be paraphrased as: find the row with the largest number of
  silver medals and then select the number of silver medals from the
  previous row.}  The spurious programs differ from the correct one in
that they repeatedly use the column ``silver''.  Whereas, in the
question, the word ``silver'' only refers to the target column
containing the answer; it also mistakenly triggers the appearance of
the column ``silver'' in the row selection condition.
This constraint, i.e.,~that a text span within a question cannot
trigger two semantically distinct operations (e.g.,~selecting target
rows and target columns)
can provide a useful inductive bias.  We propose to capture structural
constraints by modeling the alignments between programs and questions
\emph{explicitly} as structured latent variables.

Considering the large search space of possible programs, an alignment
model that takes into account the full range of correspondences
between program operations and question spans would be very expensive.
To make the process tractable, we introduce a two-stage approach that
features \emph{abstract} programs.  Specifically, we decompose
semantic parsing into two steps: 1) a natural language utterance is
first mapped to an abstract program which is a composition of
high-level operations; and 2) the abstract program is then instantiated
with low-level operations that usually involve relations and entities
specific to the knowledge base at hand.  This decomposition is
motivated by the observation that only a small number of sensible
abstract programs can be instantiated into consistent programs.
Similar ideas of using abstract meaning representations have been
explored with fully-supervised semantic parsers
\cite{dong2018coarse,data-sql-advising} and in other related tasks
\cite{goldman2018weakly,herzig-berant-2018-decoupling,nye2019learning}.

For a knowledge base in tabular format, we abstract two basic
operations of row selection and column selection from programs: these
are handled in the second (instantiation) stage.  As shown in
Figure~\ref{fig:example}, the question is first mapped to the abstract
program ``select (\#\textit{row\_slot}, \#\textit{column\_slot})"
whose two slots are subsequently instantiated with filter conditions
(row slot) and a column name (column slot).
During the instantiation of abstract programs, each slot should refer
to the question to obtain its specific semantics.  In
Figure~\ref{fig:example}, \textit{row\_slot} should attend to
``nation of Turkey" while  \textit{column\_slot} needs to attend to ``silver
medals".
The structural constraint discussed above now corresponds to assuming
that each span in a question can be aligned to a {\it unique}
row or column slot.
Under
this assumption, the instantiation of spurious programs will be
discouraged.  The uniqueness constraint would be violated by both
spurious programs in Figure~\ref{fig:example}, since ``column:silver''
appears in the program twice but can be only aligned to the span ``silver medals'' once.

The first stage (i.e.,~mapping a question onto an abstract program) is
handled with a sequence-to-sequence model.  The second stage
(i.e.,~program instantation) is approached with local classifiers: one
per slot in the abstract program.  The classifiers are conditionally
independent given the abstract program and a latent alignment.
Instead of marginalizing out alignments, which would be intractable,
we use structured attention \cite{kim2017structured}, i.e.,~we compute
the marginal probabilities for individual span-slot alignment edges
and use them to weight the input to the classifiers.  As we discuss
below, the marginals in our constrained model are computed with
dynamic programming.

We perform experiments on two
open-domain question answering datasets in the setting of learning
from denotations.  Our model achieves an execution accuracy of~44.5\%
in \wtq and 79.3\% in \wsq, which both surpass previous
state-of-the-art methods in the same weakly-supervised setting.  In
\wsq, our parser is better than recent supervised parsers that are
trained on question-program pairs.
Our contributions can be summarized as follows:
\begin{itemize}
\setlength{\parskip}{0pt}
\item we introduce an alignment model as a means of differentiating
  between correct and spurious programs;
    \item we propose a neural semantic parser that performs tractable alignments by first mapping questions to abstract programs;
    \item we achieve state-of-the-art performance on two semantic
      parsing benchmarks.\footnote{Our code is available at \url{https://github.com/berlino/weaksp_em19}.}
\end{itemize}

Although we use structured alignments to mostly enforce the uniqueness
constraint described above, other types of inductive biases can be
useful and could be encoded in our two-stage framework. For example,
we could replace the uniqueness constraint with modeling the number of
slots aligned to a span, or favor sparse alignment
distributions. Crucially, the two-stage framework makes it easier to
inject prior knowledge about datasets and formalisms while maintaining
efficiency. 


%% file: backgroud.tex
Given knowledge base~$t$, our task is to map a natural utterance~$x$
to program~$z$, which is then executed against a knowledge base to
obtain denotation~$\db{z}_t=d$.  We train our 
parser only based on~$d$ without access to correct programs~$z^*$.
Our experiments focus on two benchmarks, namely \wtq
\cite{pasupat2015compositional} and \wsq \cite{zhongSeq2SQL2017} where
each question is paired with a Wikipedia table and a denotation.
Figure~\ref{fig:example} shows a simplified example taken from \wtq.

\subsection{Grammars}\label{sec:grammar}

Executable programs~$z$ that can query tables are defined according to
a language.  Specifically, the search space of programs is constrained
by grammar rules so that it can be explored efficiently.  We adopt the
variable-free language of \citet{chen2018mapo} and define an
\emph{abstract} grammar and an \emph{instantiation} grammar which
decompose the generation of a program in two stages.\footnote{We also
  extend their grammar to additionally support operations of
  conjunction and disjunction. More details are provided in the
  Appendix.}

The first stage involves the generation of an abstract version of a
program which, in the second stage, gets instantiated.
Abstract programs only consider compositions of high-level functions, such as
superlatives and aggregation, while low-level functions and arguments,
such as filter conditions and entities, are taken into account in the next
step.  In our table-based datasets, abstract programs do not include
two basic operations of querying tables: row selection and
column selection. These operations are handled at the instantiation
stage. 
In Figure~\ref{fig:example} the abstract program has two slots for row
and column selection, which are filled with the conditions ``column:nation
= Turkey'' and ``column:silver'' at the instantiation
stage.
The two stages can be easily merged into one step when
 conducting symbolic combinatorial search. 
 The motivation for the decomposition is to facilitate the learning of 
 our neural semantic parser and the handling of structured alignments.

\paragraph{Abstract Grammar}
Our abstract grammar has five basic types: \textsc{Row},
\textsc{Column}, \textsc{string}, \textsc{number}, and \textsc{date};
\textsc{Column} is further sub-typed into \textsc{string\_column},
\textsc{number\_column}, and \textsc{date\_column}; other basic types
are augmented with \textsc{List} to represent a list of elements like
\textsc{List[Row]}.  Arguments and return values of functions are
typed using these basic types.

Function composition can be defined
recursively based on a set of production rules, each corresponding to
a function type.  For instance, function { \textsc{ROW}
  $\rightarrow$ \textit{first}(\textsc{LIST[ROW]})} selects the first
row from a list of rows and corresponds to production rule {
  ``\textsc{ROW} $\rightarrow$ \textit{first}"}. 
  
The abstract grammar
has two additional types for slots (aka terminal rules) which
correspond to row and column selection:

\begin{figure}[t]
\centering
\includegraphics[width=0.46 \textwidth]{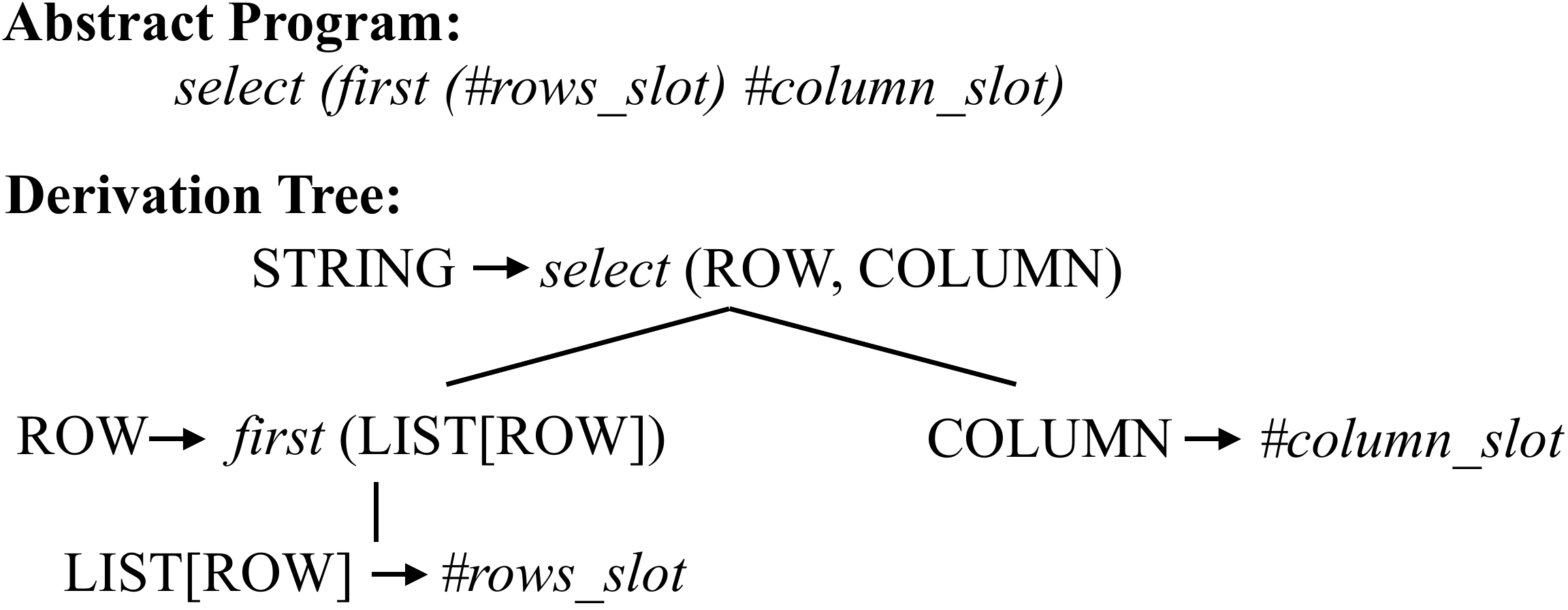}
\caption{An abstract program and its derivation tree.  Capitalized
  words indicate types of function arguments and their return value.}
\label{fig:abp}
\end{figure}



\begin{quote}
\centering
 \textsc{LIST[ROW]} $\rightarrow$ \textit{\#row\_slot} \\
 \textsc{Column} $\rightarrow$ \textit{\#column\_slot} 
\end{quote}

 An example of an abstract program and its derivation tree is shown in
 Figure~\ref{fig:abp}.  We linearize the derivation by traversing it in a left-to-right
 depth-first manner. 
 We represent the tree in
 Figure~\ref{fig:abp} as a sequence of production rules:
 ``\textsc{root} $\rightarrow$ \textsc{string}, \textsc{string}
 $\rightarrow$ \textit{select}, \textsc{row} $\rightarrow$
 \textit{first}", \textsc{list[row]} $\rightarrow$
 \textit{\#row\_slot}, \textsc{column} $\rightarrow$
 \textit{\#column\_slot}". The first action is always to select
 the return type for the root node.
 
 Given a specific table~$t$, the abstract grammar~$\mathcal H_t$ will
 depend on its column types.  For example, if the table does not have
 number cells, ``max/min'' operations   will not be executable.


\paragraph{Instantiation Grammar}
A column slot is directly instantiated by selecting a column; a row
slot is filled with one or multiple conditions (\textsc{cond}) which are joined
together with conjunction ({\small \textsl{OR}}) and disjunction
({\small \textsl{AND}}) operators:
\begin{quote}
\textsc{cond} $\rightarrow$ \textsc{column} \textsc{operator}  \textsc{value}  \\
   \textit{\#row\_slot} $\rightarrow$  \textsc{cond} ({\small \textsl{AND}}  \textsc{cond})* \\
   \textit{\#row\_slot} $\rightarrow$ \textsc{cond} ({\small  \textsl{OR}}  \textsc{cond})* \\
\end{quote}
\vspace{-1.5mm}
 where { \textsc{Operator}} $\in [>, <, =, \geq, \leq]$ and
 {\textsc{value}} is a string, a
 number, or a date.  A special condition {
   $\textit{\#row\_slot} \rightarrow all\_rows$ } is defined to signify that a
 program queries all rows.

\subsection{Search for Consistent Programs}
\label{sec:coverage}

A problematic aspect of learning from denotations is that, since
annotated programs are not available (e.g., for~\wtq), we have no
means to directly evaluate a proposed
grammar. 
%
As an evaluation proxy, we measure the
coverage of our grammar in terms of consistent programs. Specifically,
we exhaustively search for all consistent programs for each question
in the training set.
While the space of programs is exponential, we observed that abstract programs which are
instantiated into correct programs are not very complex in terms of the number of production rules used to generate them. 
As a result, we impose restrictions on  the number  of production rules which can abstract programs, and in this way the search process becomes tractable.\footnote{Details are provided in the Appendix.}


We find that 83.6\% of questions in \wtq are covered by at least one
consistent program.  However, each question eventually has 200
consistent programs on average and most of them are spurious.
Treating them as ground truth poses a great challenge for learning a
semantic parser.  The coverage for \wsq is 96.6\% and each question
generates 84~consistent programs.

Another important observation is that there is only a limited number
of abstract programs that can be instantiated into consistent
programs.  The number of such abstract programs is~23 for \wtq and~6
for \wsq, suggesting that there are a few patterns underlying several
utterances.  This motivates us to design a semantic parser that first
maps utterances to abstract
programs. 
For the sake of generality, we do not restrict our parser to abstract
programs in the training set. We elaborate on this below.


%% file: model.tex
After obtaining consistent programs~$z$ for each question via
offline search, we next show how to learn a parser that can generalize
to unseen questions and tables.

\subsection{Training and Inference}

Our learning objective $\mathcal J$
is to maximize the log-likelihood of the marginal probability of all
consistent programs, which are generated by mapping an utterance~$x$
to an interim abstract program~$h$:
\begin{equation}
\mathcal J =\log \{ \sum\limits_{h\in \mathcal{H}_t} {p(h|x,t)\sum\limits_{\db{z} = d} {p(z|x,t,h)} } \}.
\end{equation}

During training, our model only needs to focus on abstract programs
that have successful instantiations of consistent programs
and it does not have to explore the whole space of possible programs.

At test time, the parser chooses the program $\hat z$ with the highest probability:
\begin{equation}
\hat h, \hat z =\argmax_{h\in \mathcal{H}_t, \ z}  {p(h|x,t) p(z|x,t,h)}.
\end{equation}
For efficiency, we only choose the \mbox{top-$k$} abstract programs to instantiate through beam search.
$\hat z$ is then executed to obtain its denotation as the final prediction.

Next, we will explain the basic components of our neural parser.
Basically, our model first encodes a question and a table with an
input encoder; it then generates abstract programs with a seq2seq
model; and finally, these abstract programs are instantiated based on a structured alignment model.

\subsection{Input Encoder}

Each word in an utterance is mapped to a distributed representation
through an embedding layer.  Following previous work
\cite{neelakantan2016learning,chen2018mapo}, we also add an indicator
feature specifying whether the word appears in the table. This feature is
mapped to a learnable vector.  Additionally, in \wtq, we use POS tags
from the CoreNLP annotations released with the dataset and map them to
vector representations.  The final representation for a word is the
concatenation of the vectors above.  A bidirectional LSTM
\cite{hochreiter1997long} is then used to obtain a contextual
representation $\vl_i$ for the~$i_{th}$ word.

A table is represented by a set of columns.
Each column is encoded by averaging the embeddings of words under its column name.
We also have a column type feature (i.e., number, date, or string) and
an indicator feature signaling whether at least one entity in the column appears in the utterance.


\subsection{Generating Abstract Programs}

Instead of extracting abstract programs as templates, similarly to
\citet{xu2017sqlnet} and \citet{data-sql-advising}, we generate them
with a seq2seq model. Although template-based approaches
  would be more efficient in practice, a seq2seq model is more general
  since it could generate unseen abstract programs which fixed
  templates could not otherwise handle.

 Our goal here is to generate
a sequence of production rules that lead to abstract programs.  During
decoding, the hidden state $\vg_j$ of the~$j$-th timestep is
computed based on the previous production rule, which is mapped to an
embedding~$\va_{j-1}$.  We also incorporate an attention mechanism
\cite{luong2015effective} to compute a contextual vector $\vb_j$.
Finally, a score vector $\vs_j$ is computed by feeding the
concatenation of the hidden state and context vector to a multilayer
perceptron (MLP):
\begin{align}
\begin{split}
\vg_j & = \LSTM (\vg_{j-1}, \va_{j-1}) \\
\vb_j & = \att (\vg_j, \vl) \\
\vs_j & =  \mathop{\rm MLP_1}  ([\vg_j; \vb_j]) \\
p(a_j|x,t,a_{<j}) &= \softmax_{a_j} (\vs_j)
\end{split}
\end{align}
%
where the probability of production rule~$a_j$ is computed by the
softmax function.  According to our abstract grammar, only a subset of
production rules will be valid at the~$j$-{th} time step.  For
instance, in Figure~\ref{fig:abp}, production rule ``\textsc{string}
$\rightarrow$ \textit{select}" will only expand to rules whose
left-hand side is \textsc{row}, which is the type of the first
argument of \textit{select}.  In this case, the next production rule
is ``\textsc{row} $\rightarrow$ \textit{first}".  We thus restrict
the normalization of softmax to only focus on these valid production
rules.
The probability of generating an abstract program~$p(h|x,t)$ is simply
the product of the probability of predicting each production rule
$\prod_j p(a_j|x,t,a_{<j})$.

After an abstract program is generated,  we need to instantiate slots in abstract programs.
Our model first encodes the abstract program using a bi-directional LSTM.
As a result, the representation of a slot is contextually aware
of the entire abstract program \cite{dong2018coarse}.

\subsection{Instantiating Abstract Programs}

To instantiate an abstract program, each slot must obtain its specific
semantics from the question.  We model this process by an alignment
model which learns the correspondence between slots and question
spans.  Formally, we use a binary alignment matrix $\mA$ with size
${m \times n \times n}$, where $m$ is the number of slots and $n$ is
the number of tokens.
In Figure~\ref{fig:example}, the alignment matrix will only have
$\mA_{0,6,8}=1$ and $\mA_{1,2,3}=1$ which indicates that the first slot
is aligned with ``nation of Turkey'', and the second
slot is aligned with ``silver medals''.
The second and third dimension of the matrix represent the start and end position of a span.

We model alignments as discrete latent variables and condition the
instantiation process on the alignments as follows:
\begin{equation}
\begin{array}{ll}
\sum\limits_{[z]  = d} p(z|x,t,h) = & \\ \hspace*{.65cm}{\sum\limits_\mA {p(\mA |x,t,h)\sum\limits_{[z] = d} {p(z|x,t,h,\mA)} } }.
\end{array}
\label{eq:align}
\end{equation}

We will first discuss the instantiation model ${p(z|x,t,h,\mA)}$ and
then elaborate on how to avoid marginalization in the next section.
Each slot in an abstract program can be instantiated by a set of
candidates following the instantiation grammar.  For efficiency, we
use local classifiers to model the instantiation of each slot
independently:
\begin{equation}
p(z|x, t, h,\mA) = \prod_{s\in S} p(s \rightarrow c|x, t, h, \mA),
\end{equation}
%
where $S$ is the set of slots and $c$ is a candidate following our
instantiation grammar.
``$s \rightarrow c$'' represents the instantiation of slot $s$ into candidate $c$.

Recall that there are two types of slots, one for rows and one for
columns.  All column names in the table are potential instantiations
of column slots. We represent each column slot candidate by the
 average of the embeddings of words in the column name.
Based on our instantiation grammar in Section \ref{sec:grammar}, candidates for row slots are
represented as follows: 1) each condition is represented with the
concatenation of the representations of a column, an operator, and a
value.  For instance, condition ``string\_column:nation = Turkey'' in
Figure~\ref{fig:example} is represented by vector representations of
the column `nation', the operator `=', and the entity `Turkey'; 2)
multiple conditions are encoded by averaging the representations of
all conditions and adding a vector representation of {\small
  \textsl{AND}}/{\small \textsl{OR}} to indicate the relation between them.

For each slot, the probability of generating a candidate is computed with
softmax normalization on a score function:
\begin{equation}
 p(s \rightarrow c|x,t,h,\mA) \propto  \exp \{ \mathop{\rm MLP}  ([\vs; \vc]) \},
 \label{eq:slot_inst}
 \end{equation}
%
 where $\vs$ is the representation of the span that slot~$s$ is aligned with,
 and $\vc$ is the representation of candidate $c$.  The
 representations $\vs$ and $\vc$ are concatenated and fed to
 a MLP. We use the same MLP architecture but different
 parameters for column and row slots.

\subsection{Structured Attention}
\label{sect:struct-attent}


We first formally define a few structural constraints over alignments
and then explain how to incorporate them efficiently into our parser.

The intuition behind our alignment model is that row and column
selection operations represent distinct semantics,
and should therefore be expressed by distinct natural language
expressions.  Hence, we propose the following constraints:

\paragraph{Unique Span}
In most cases, the semantics of a row selection or a column selection
is expressed uniquely with a single contiguous span:
\begin{equation}
\forall k \in [1, |S|], \quad \sum_{i,j} \mA_{k,i,j}  = 1,
\end{equation}
%
where $|S|$ is the number of slots.

\paragraph{No Overlap}
Spans aligned to different slots should not overlap.  Formally, at
most one span that contains word~$i$ can be aligned to a slot:
\begin{equation}
\forall i \in [1, n], \quad \sum_{k,j} \mA_{k,i,j}  \leq 1.
\end{equation}

As an example, the alignments in Figure~\ref{fig:example} follow the
above constraints.  Intuitively, the one-to-one mapping constraint
aims to assign distinct and non-overlapping spans to slots of abstract
programs.
To further bias the alignments and improve efficiency, we impose additional restrictions:
 (1) a row slot must be aligned to a span that contains an
entity since conditions that instantiate the slot would require
entities for filtering; (2) a column slot must be aligned to a span with
length~1 since most column names only have one word.

Marginalizing out all $\mA$ in Equation~(\ref{eq:align}) would be very
expensive considering the exponential number of possible alignments.
We approximate the marginalization by moving the outside expectation
directly inside over $\mA$.  As a result, we instead optimize the
following objective:
\begin{equation}
\small{
\mathcal J \approx \log \big\{ \sum\limits_{h \in \mathcal{H}_t} {p(h |x,t)} \sum\limits_{\db{z} = d} {p(z|x, t, h, \E [\mA])}  \big\}
},
\end{equation}
where $\E[\mA]$ are the marginals of $\mA$ with respect to $p(\mA|x,t,h)$.

 The idea of using differentiable surrogates for discrete latent variables has been used
 in many other works like differentiable data structures \cite{grefenstette2015learning,graves2014neural}
 and attention-based networks \cite{bah15,kim2017structured}.
Using marginals
$\E[\mA]$ can be viewed
as \textit{structured attention} between slots and question spans.

The marginal probability of the alignment matrix $\mA$ can be computed
efficiently using dynamic programming (see \citealt{dp2015srl} for
details).  An alignment is encoded into a path in a weighted lattice
where each vertex has $2^{|S|}$ states to keep track of the set of covered slots.
The marginal probability of edges in this lattice can be
computed by the forward-backward
algorithm~\cite{wainwright2008graphical}.  The lattice weights,
represented by a scoring matrix $\mM \in \R^{m \times n \times n}$ for
all possible slot-span pairs, are computed using the following scoring
function: 
\begin{equation}
\mM_{k,i,j} =  \mathop{\rm MLP_2}  ([\vr(k); {\bf span}[i:j]]),
 \end{equation}
 where $\vr(k)$ represents the $k_{th}$ slot and ${\bf span}[i:j]$ represents the
 span from word~$i$ to~$j$.
 Recall that we obtain ~$\vr(k)$ by encoding a generated abstract
 program.  A span is represented by averaging the
 representations of the words therein.
 These two representations are concatenated and fed to a $ \mathop{\rm MLP}$ to obtain a score.
 Since $\E[\mA]$ is not discrete anymore, the aligned representation of slot $\vs$ in
 Equation~(\ref{eq:slot_inst}) becomes the weighted average of representations of all spans in the set.


%% file: experiment.tex
We evaluated our model on two semantic parsing benchmarks, \wtq and
\wsq.  We compare  against two common baselines to
demonstrate the effectiveness of using abstract programs and
alignment. We also
conduct detailed analysis which shows that structured attention is
highly beneficial, enabling our parser to differentiate between correct
and spurious programs.  Finally, we break down the errors of our
parser so as to examine whether structured attention is better at
instantiating abstract programs.

\subsection{Experimental Setup}

\paragraph{Datasets} \wtq contains 2,018 tables and 18,496
utterance-denotation pairs.  The dataset is challenging as 1)
the tables cover a wide range of domains and unseen tables appear at
test time; and 2) the questions involve a variety of operations such as
superlatives, comparisons, and aggregation
\cite{pasupat2015compositional}.  \wsq has 24,241 tables and 80,654
utterance-denotation pairs.  The questions are logically simpler and
only involve aggregation, column selection, and conditions.  The
original dataset is annotated with SQL queries, but we only use the
execution result for training.  In both datasets, tables are
extracted from Wikipedia and cover a wide range of domains.

Entity extraction is important during parsing since entities are used
as values in filter conditions during instantiation.  String entities
are extracted by string matching utterance spans and table
cells.  In \wtq, numbers and dates are extracted from the CoreNLP
annotations released with the dataset.  \wsq does not have entities
for dates, and we use string-based normalization to deal with numbers.

\paragraph{Implementation} We obtained word embeddings by a linear
projection of GloVe pre-trained embeddings \cite{pennington2014glove}
which were fixed during training.  Attention scores were computed
based on the dot product between two vectors.  Each $\mathop{\rm MLP}$
is a one-hidden-layer perceptron with ReLU as the activation function.
Dropout \cite{srivastava2014dropout} was applied to prevent
overfitting. All models were trained with Adam
\cite{kingma2014adam}. Implementations of abstract and instantiation grammars were based
on AllenNLP \cite{Gardner2017AllenNLP}.\footnote{Please refer to the Appendix for the full list of hyperparameters used in our experiments.}

\begin{table}[t]
\centering
  \begin{tabular}{lccc}
  	\toprule
    \textbf{Supervised by Denotations}  &  Dev. & Test \\
    \midrule
 \citet{pasupat2015compositional} & 37.0 & 37.1 \\
 \citet{neelakantan2016learning} & 34.1 & 34.2 \\
 \citet{haug2017neural} & --- & 34.8 \\
 \citet{zhang-etal-2017-macro} & 40.4 & 43.7 \\
\citet{chen2018mapo} & 42.3 & 43.1  \\
\citet{dasigi-etal-2019-iterative} & 42.1 & 43.9 \\
\citet{agarwal2019learning}& 43.2 & 44.1  \\
\midrule
Typed Seq2Seq & 37.3 & 38.3 \\
Abstract Programs \\
\quad \textit{f.w.} standard attention & 39.4  & 41.4 \\
\quad \textit{f.w.}  structured attention & \textbf{43.7} & \textbf{44.5} \\
    \bottomrule
  \end{tabular}
\caption{Results on \wtq.
\textit{f.w.}~stands for  slots filled with.
}
\label{tab:wtq}
\end{table}

\subsection{Baselines}
Aside from comparing our model against previously published
approaches, we also implemented the following baselines:

\paragraph{Typed Seq2Seq} Programs were generated using a
sequence-to-sequence model with attention \cite{dong2016language}.
Similarly to \citet{krishnamurthy-etal-2017-neural}, we constrained the
decoding process so that only well-formed programs are predicted.
This baseline can be viewed as merging the two stages of our model
into one stage where generation of abstract programs and their
instantiations are performed with a shared decoder.

\paragraph{Standard Attention}
The aligned representation of slot $s$ in
Equation~(\ref{eq:slot_inst}) is computed by a standard attention
mechanism: $\vs = \att (\vr(s), \vl) $ where $\vr(s)$ is the
representation of slot $s$ from abstract programs.  Each slot is
aligned independently with attention, and there are no global structural
constraints on alignments. 

\subsection{Main Results}

For all experiments, we report the mean accuracy of 5 runs.
Results on \wtq are shown in Table \ref{tab:wtq}.  The
structured-attention model achieves the best performance, compared
against the two baselines and previous approaches.  The standard
attention baseline with abstract programs is superior to the typed
Seq2Seq model, demonstrating the effectiveness of decomposing semantic
parsing into two stages.
Results on \wsq are shown in Table \ref{tab:wsq}.  The
structured-attention model is again superior to our two baseline
models.  Interestingly, its performance surpasses previously reported
weakly-supervised models \cite{chen2018mapo,agarwal2019learning} and
is on par even with fully supervised ones \cite{dong2018coarse}.

The gap between the standard attention baseline and the typed Seq2Seq
model is not very large on \wsq, compared to \wtq.
Recall from Section~\ref{sec:coverage} that \wsq only has 6~abstract programs
that can be successfully instantiated. For this reason, our
decomposition alone may not be very beneficial if coupled with standard attention.
In contrast, our structured-attention model consistently performs much better than both baselines.

We report scores of ensemble
systems in  Table~\ref{tab:ens}.
We use the best model which relies on abstract programs
and structured attention as a base model in our ensemble.
Our ensemble system achieves better performance than \citet{chen2018mapo} and \citet{agarwal2019learning}, while using the same ensemble size.

\begin{table}[t]
\centering
  \begin{tabular}{lccc}
  	\toprule
    \textbf{Supervised by Programs} &  Dev. & Test \\
    \midrule
 \citet{zhongSeq2SQL2017}      &  60.8 & 59.4 \\
\citet{chenglong} &67.1 & 66.8 \\
\citet{xu2017sqlnet}  & 69.8 & 68.0 \\
\citet{pshuang2018PT-MAML} & 68.3 & 68.0 \\
\citet{yu2018typesql} & 74.5 & 73.5 \\
\citet{sun2018semantic} & 75.1 & 74.6 \\
\citet{dong2018coarse}  & {79.0} & {78.5} \\
\citet{shi2018incsql} & \textbf{84.0} & \textbf{83.7} \\
    \midrule
    \midrule
    \textbf{Supervised by Denotations} & Dev. & Test \\
    \midrule
\citet{chen2018mapo} & 72.2 & 72.1  \\
\citet{agarwal2019learning}& 74.9 & 74.8  \\
\midrule
Typed Seq2Seq & 74.5 & 74.7 \\
Abstract Programs \\
\quad \textit{f.w.} standard attention & 75.2 & 75.3 \\
\quad \textit{f.w.}  structured attention & 79.4 & 79.3 \\
    \bottomrule
  \end{tabular}
\caption{Results on \wsq.
\textit{f.w.}:  slots filled with.
}
\label{tab:wsq}
\end{table}

\subsection{Analysis of Spuriousness}

To understand how well structured attention can help a parser
differentiate between correct and spurious programs, we analyzed the
posterior distribution of consistent programs given a denotation:
$p(z|x,t,d)$ where $\db{z}=d$.

\wsq includes gold-standard SQL annotations, which we do not use in
our experiments but exploit here for analysis.  Specifically,
we converted the annotations released with \wsq to programs licensed
by our grammar.  We then computed the log-probability of these
programs according to the posterior distribution as a measure of how
well a parser can identify them amongst all consistent programs
$\log \sum_{z^*} p(z^*|x,t,d)$, where $z^*$ denotes correct programs.
The average log-probability assigned to correct programs by structured
and standard attention is -0.37 and -0.85, respectively.  This gap
confirms that structured attention can bias our parser towards correct
programs during learning.

\begin{table}
\centering
  \begin{tabular}{lcccc}
  	\toprule
    \textbf{Models} &  \swtq & \wsq \\
    \midrule
\citet{chen2018mapo} &  46.3  (10) &  74.2 (5)\\
\citet{agarwal2019learning} & 46.9 (10)   & 76.9 (5)\\
Our model  & \textbf{47.3} (10)   & \textbf{81.7} (5) \\
    \bottomrule
  \end{tabular}
\caption{ Results of ensembled models on the test set; ensemble sizes are shown within parentheses.}
\label{tab:ens}
\vspace{3mm}
\end{table}

\subsection{Error Analysis}

\begin{table}
\centering
  \begin{tabular}{lcccc}
  	\toprule
    \textbf{Error Types} &  standard & structured \\
    \midrule
Abstraction Error  & 19.2   & 20.0 \\
Instantiation Error & 41.5  & 36.2 \\
Coverage Error  & 39.2  & 43.8 \\
    \bottomrule
  \end{tabular}
\caption{Proportion of errors on the development set in \wtq.}
\label{tab:error}
\end{table}

We further manually inspected the output of our structured-attention
model and the standard attention baseline in \wtq.
Specifically, we randomly sampled 130 error cases independently from
both models and  classified them into three categories.

\paragraph{Abstraction Errors}
If a parser fails to generate an abstract program, then it is
impossible for it to instantiate a consistent complete program.

\paragraph{Instantiation Errors}
These errors arise when abstract programs are correctly generated, but
are mistakenly instantiated either by incorrect column names or filter
conditions.

\paragraph{Coverage Errors}
These errors arise from implicit assumptions made by our parser: a)
there is a long tail of unsupported operations that are not covered by
our abstract programs; b) if entities are not correctly identified and
linked, abstract programs cannot be correctly instantiated.

Table~\ref{tab:error} shows the proportion of errors attested by the
two attention models.
We observe that structured attention suffers less from instantiation
errors compared against the standard attention baseline, which points
to the benefits of the structured alignment model.

%% file: related_work.tex
\paragraph{Neural Semantic Parsing}

We follow the line of work that applies sequence-to-sequence models
\cite{sutskever2014sequence} to semantic parsing
\cite{jia2016data,dong2016language}. Our work also relates to models
which enforce type constraints \cite{yin2017syntactic,
  rabinovich2017abstract,krishnamurthy-etal-2017-neural} so as to
restrict the vast search space of potential programs.  We use both
methods as baselines to show that the structured bias introduced by
our model can help our parser handle spurious programs in the setting
of learning from denotations.  Note that our alignment model can also
be applied in the supervised case in order to help the parser rule
out incorrect programs.

Earlier work has used lexicon mappings
\cite{zettlemoyer2007online,wong2007learning,lu2008generative,kwiatkowski2010inducing}
to model correspondences between programs and natural
language. However, these methods cannot generalize to unseen tables
where new relations and entities appear.
To address this issue, \citet{pasupat2015compositional} propose a
floating parser which allows partial programs to be generated without
being anchored to question tokens.  In the same spirit, we use a
sequence-to-sequence model to generate abstract programs while relying
on explicit alignments to instantiate them.
Besides semantic parsing, treating alignments as discrete latent variables has proved effective in other tasks like sequence transduction \cite{yu-etal-2016-online} and AMR parsing \cite{lyu2018amr}.


\paragraph{Learning from Denotations}

To improve the efficiency of searching for consistent programs,
\citet{zhang-etal-2017-macro} use a macro grammar induced from cached
consistent programs.  Unlike \citet{zhang-etal-2017-macro} who
abstract entities and relations from logical forms, we take a step
further and abstract the computation of row and column selection.  Our
work also differs from \citet{pasupat2016inferring} who resort to
manual annotations to alleviate spuriousness. Instead, we equip our
parser with an inductive bias to rule out spurious programs during
training.  Recently, reinforcement learning based methods address the
computational challenge by using a memory buffer \cite{chen2018mapo} which
stores consistent programs and an auxiliary reward function
\cite{agarwal2019learning} which provides feedback to deal with
spurious programs.  \citet{guu2017language} employ various strategies
to encourage even distributions over consistent programs in cases
where the parser has been misled by spurious programs.
\citet{dasigi-etal-2019-iterative} use coverage of lexicon-like rules to guide the search of consistent programs.


%% file: conclusion.tex
In this paper, we proposed a neural semantic parser that learns from
denotations using abstract programs and latent structured alignments.
Our parser achieves state-of-the-art performance on two benchmarks,
\wtq and \wsq.  Empirical analysis shows that the inductive bias
introduced by the alignment model helps our parser differentiate
between correct and spurious programs. Alignments can exhibit
different properties (e.g., monotonicity or bijectivity), depending on
the meaning representation language (e.g., logical forms or SQL), the
definition of abstract programs, and the domain at hand.  We believe
that these properties can be often captured within a probabilistic
alignment model and hence provide a useful inductive bias to the
parser.



%% file: appendix.tex
\section{Grammars}

We created our grammars following \citet{zhang-etal-2017-macro} and  \citet{chen2018mapo}.
Compared with \citet{chen2018mapo}, we additionally support disjunction({\small \textsl{OR}})  and conjunction({\small \textsl{AND}}).
Some functions are pruned based on their effect on coverage, which is the proportion of questions that obtain at least one consistent program.
``same\_as" function \cite{chen2018mapo} is excluded since it introduces too many spurious programs while contributing little to the coverage.
For the same reason, conjunction({\small \textsl{AND}}) is not used in \wtq and disjunction({\small \textsl{OR}}) is not used in \wsq.

We also include non-terminals of function types in production rules \cite{krishnamurthy-etal-2017-neural}.
For instance,   function ``{\small \textsc{ROW} $\rightarrow$ \textit{first}(\textsc{LIST[ROW]})}" selects the first row from a list of rows and 
will lead to  the production rule  {\small ``\textsc{ROW} $\rightarrow$ \textit{first}"}.
In the paper, we eliminate the function type for simplicity.
Practically, we use two production rules to represent the function: {\scriptsize \textsc{ROW} $\rightarrow$ $<$\textsc{ROW}: \textsc{LIST[ROW]} $>$} and {\scriptsize $<$\textsc{ROW}:\textsc{LIST[ROW]}$>$ $\rightarrow$ \textit{first }},
where {\scriptsize $<$ \textsc{ROW}: \textsc{LIST[ROW]} $>$} is an abstract function type. 

\section{Search for Consistent Programs}

We enumerate all possible programs in two stages using the abstract and instantiation grammars. 
To constrain the space and make the search process tractable, we restrict the maximal number of production rules for generating abstract programs during the first stage. 
It is based on the observation that the abstract programs which can be successfully instantiated into \textit{correct} programs are usually not very complex.
In other words, the \textit{consistent} programs that are instantiated by long abstract programs are very likely to be spurious. 
For instance, programs like ``{\small select( previous( next( previous(argmax [all\_rows], column:silver) column:bronze)}" are unlikely to have a corresponding question.  
Specifically, we set the maximal number of production rules for generating abstract programs to  6 and 9, which leads to search time of around 7 and 10 hours for \wsq and \wtq respectively, using a single CPU. 
Note that this needs to be done only once.

\section{Hyperparameters}

Models used in \wtq and \wsq share similar hyperparameters which are listed in Table~\ref{tab:hyper}. 
Our input embeddings are obtained by a linear projection from the fixed pre-trained embedding \cite{pennington2014glove}. 
Word Indicator refers to the indicator feature of whether a word appears in the table;
Column Indicator refers to the indicator feature of whether at least one entity in a column appears in the question.
All MLPs mentioned in the paper have the same hidden size and dropout rate.
During decoding, we choose the top-6 abstract programs to instantiate via beam search.

\begin{table}[t]
\centering 
\small
  \begin{tabular}{lcccc}
  \toprule
Hyperparameters &  \swtq & \wsq \\
    \midrule
Input Fixed Embedding Size & 300  & 300 \\
Input  Linear Projection Size & 256  & 256\\
POS Embedding Size & 64 & - \\
Production Rule Embedding Size& 436  & 328 \\
Column Type Embedding Size & 16 & 16 \\
Word Indictor Embedding Size & 16 & 16 \\
Column Indictor Embedding Size & 16 & 16 \\
Operator Embedding Size & 128  & 128 \\
Encoder Hidden Size & 256  & 256 \\
Encoder Dropout & 0.45  & 0.35 \\
Decoder Hidden Size & 218  & 164 \\
AP Encoder Hidden Size  & 218  & 164 \\
AP Encoder Dropout  & 0.25  & 0.25 \\
MLP Hidden Size & 436  & 328 \\
MLP Dropout & 0.25  & 0.2 \\
    \bottomrule
  \end{tabular}
\caption{Hyperparameters for \swtq~(\wtq) and \wsq. AP Encoder is the encoder representing  the abstract programs we generate.}.
\label{tab:hyper}
\end{table}

\section{Alignments}

If a row slot is instantiated with the special condition `all\_rows', 
then it is possible that the semantics of this slot is implicit.
For instance, the question ``which driver completed the least number of laps? " should first be mapped to the abstract program 
``{\small select (argmin (\#\textit{row\_slot}, \#\textit{column\_slot}), \#\textit{column\_slot} )}" which is then instantiated to 
the correct program ``{\small select (argmin (all\_rows, column:laps) column:driver)}".
The row slot in the abstract program is instantiated with `all\_rows', but this is not explicitly expressed in the question.
 
 To make it compatible with our constraints in Equation (6) and (7), 
 we add a special token ``ALL\_ROW'' at the end of each question.
 If a row slot is aligned with this token, 
 then it is expected to be instantiated with the special condition `all\_rows'. 
 Specifically, this special token is mapped to a learnable vector during instantiations.
Our alignment needs to learn to align this special token with a row slot if this row slot should be instantiated with the condition `all\_rows'.